\documentclass[12pt]{article}
\usepackage{amsmath}
\usepackage{times}
\usepackage{graphicx}
\usepackage{color}
\usepackage{multirow}
\usepackage{rotating}
\usepackage{bbm}
\usepackage{latexsym}

\makeatletter  
\long\def\@makecaption#1#2{%
  \vskip\abovecaptionskip
  \sbox\@tempboxa{{\captionfonts #1: #2}}%
  \ifdim \wd\@tempboxa >\hsize
    {\captionfonts #1: #2\par}
  \else
    \hbox to\hsize{\hfil\box\@tempboxa\hfil}%
  \fi
  \vskip\belowcaptionskip}
\makeatother   

\usepackage{amssymb,comment}

\usepackage{nameref}

\usepackage[table]{xcolor}

\usepackage{amsthm}
\usepackage{amsfonts}
\usepackage{amssymb}
\usepackage{graphicx}
\usepackage{mathalfa}
\usepackage{psfrag}
\usepackage[justification=centerfirst]{subfig}

\newcommand{\one}{($i$) }
\newcommand{\two}{($ii$) }
\newcommand{\three}{($iii$) }
\newcommand{\four}{($iv$) }

\usepackage[utf8]{inputenc}
\usepackage{verbatim}
\usepackage{tikz}
\usepackage{wrapfig}
\usepackage[authoryear]{natbib}
\usepackage{caption}
\usepackage{float}

\begin{document}
\hspace{13.9cm}1

\ \vspace{20mm}\\

{\LARGE Minimal spiking neuron for solving multi-label classification tasks}

\ \\
{\bf \large Jakub Fil and Dominique Chu}\\
{School of Computing, University of Kent, CT2 7NF, Canterbury, United Kingdom \\ \{j.fil,dfc\}@kent.ac.uk}\\
%


\thispagestyle{empty}
\markboth{}{NC instructions}
%
%
\begin{center} {\bf Abstract} \end{center}
The Multi-Spike Tempotron (MST) is a powerful single spiking neuron model that can solve complex supervised classification tasks. While powerful, it is also internally complex, computationally expensive to evaluate, and not suitable for neuromorphic hardware. Here we aim to understand whether it is possible to simplify the MST model, while retaining its ability to learn and to process information. To this end,  we introduce a family of Generalised Neuron Models (GNM) which are a special case of the Spike Response Model and much simpler and cheaper to simulate than the MST. We find  that over a wide range of parameters the GNM can learn at least as well as the MST. We identify the temporal autocorrelation of the membrane potential as the single most important ingredient of the GNM which enables it to classify multiple spatio-temporal patterns. We also interpret the GNM as a chemical system,  thus conceptually bridging computation by neural networks  with molecular information processing.  We conclude the paper by proposing alternative training approaches for the GNM including error trace learning and error backpropagation.  
\section{Introduction}

Spiking neurons have been shown to be computationally more powerful than standard rate coded neurons. \citep{maass1997, Gutig2006, Rubin2010}. This motivates the hope that networks of spiking neurons (SNNs) can be built smaller than corresponding networks of rate-coded neurons, while achieving the same computational task. Practically, this would be interesting because current deep architecture, while solving sophisticated tasks, also require extremely large models. Networks requiring up to a billion of weights are not uncommon, for example in language processing \citep{Radford2018} or image recognition \citep{Mahajan2018}. 
\par
In practice, there are difficulties, however. Spiking neurons are more expensive to simulate than non-spiking units. This may outweigh any gains from the reduced network size. In some cases, this problem may be circumvented by using specialised neuromorphic hardware \citep{Giacomo2011, Plana2011,Lin2018,memory_spike1, memory_spike2} to model SNNs. However, this is not always practical or possible  \citep{BrainScaleS}.
\par
In most cases it will remain necessary to simulate SNNs on general purpose computers. In order to be able to build efficient SNNs, we therefore need to understand systematically the computational properties of spiking neurons.  Indeed, there exists  a large number of variants of spiking models vastly varying in internal complexity \citep{hodgkin_huxley_1952, izhikevich2003, Brunel2007, Gerstner2014}. These various neuronal models are not always developed with computationally efficiency as a criterion. Particularly in computational neuroscience considerations of (the rather vague concept of)  ``biological plausibility'' are often more important than pure simplicity. Yet, in the context of applications of spiking neurons in AI, biological plausibility is  irrelevant and the important criteria should be computational cost, ease of implementation, suitability for neuromorphic hardware, and performance. In order to be able to produce maximally parsimonious models it is essential to first understand the necessary ingredients that enable computation. This will be the focus of this contribution.
\par
Concretely, here we will investigate the minimal ingredients required for a single neuron to perform a multi-label classification task. The purpose of this task is to classify incoming spatio-temporal patterns into different classes and to distinguish them from noise. In the SNN literature, there have been a number of attempts to solve variants of this task, including the   {\em Remote Supervised Method} (ReSuMe)  \cite{Ponulak2005}, the Chronotron learning rule \citep{Florian2012}  or the   {\em Spike Pattern Association Neuron}  \citep{MOHEMMED2012} and the {\em Precise Spike Driven Synaptic Plasticity}  \citep{Yu2013}. For the purpose of this article, we will focus on  one of the most recent approaches  --- the   {\em Multi-Spike Tempotron} (MST) \citep{Gutig2016}.  This is a single neuron architecture that can be trained to distinguish fixed spatio-temporal patterns from (statistically indistinguishable) noise.  Crucially, the MST can also classify patterns into different classes, where the class label is indicated by the   number of output spikes released during the duration of the pattern. Noise is always considered as class ``0'', i.e. the MST should not spike when presented with noise. 
\par
The neural dynamics of the MST can be summarised as follows: \one The state of the neuron is defined by the value of the internal ``membrane potential'' $V(t)$. It is updated in discrete time. Spikes are generated when $V(t)$ crosses a set threshold value from below.  \two The input to the MST neuron are $N$ (unmodelled) ``pre-synaptic'' neurons, spiking with a set frequency. \three The MST does not accept directly spiking input, but includes a preprocessing step whereby input spikes are converted into  analog signals via a  bi-exponential synapse. This could be interpreted as an additional layer of trivial, capacitor-like buffer neurons. \four  The membrane potential update rule of the MST takes into account the total spiking history of the pre-synaptic neurons. As a result the future evolution of the neuron does not solely depend on the current state and inputs, but it also depends on how the current state was reached.   The update function also includes  a constant decay of the membrane potential and a ``soft'' exponential reset following a spike. Unlike, for example, the leaky integrate and fire (LIF) neuron, the MST does not have an immediate ``hard'' reset to the resting potential nor a refractory period.  
\par
In summary, while the MST is a powerful neuronal model, it is also internally complex. The question we wish to address here is whether or not this internal complexity is necessary for the ability of the MST to learn. We shall find that it is not. Much simpler models can learn at least equally well.  To show this,  we  shall systematically  strip away features from the MST and check whether this impacts on its ability to learn.  To this end, we  propose a family of {\em  generalised neuronal models} (GNM), which is a special case of the well known Spike-Response Model \citep{Gerstner2002, Jolivet_SRM}. The GNM contains a number of   readily interpretable parameters which we vary systematical in order to explore putatively crucial features of the model. Depending on how the parameters are set, we can approximate the MST model or implement radically simpler models. The most important parameters that we shall find are  the ``spikiness''  of the GNM and its ``memory''. 
\par
Using a rigorous exploration of the GNM parameter space, we find that most of the complexities of the MST neuron are not essential for learning. Indeed, there is no strict need for spiking, nor is the soft-reset important. However, we do find that  a balanced amount of memory of past states, i.e. a degree of temporal autocorrelation of the membrane potential, is crucial for learning. Interestingly, we shall identify the hard reset of the well known leaky integrate and fire neuron (LIF) \citep{Gerstner2014}, which erases any memory of pre-spike states, destroying the correlation of post-spike and pre-spike membrane potentials, as a hindrance to learning in the single neuron model. 
\par
Following common practice in SNN, we assume that the GNM is updated in discrete time. However, we also find that a continuous time version of the GNM can classify well. This continuous time version is theoretically interesting because it  lends itself to an interpretation as a chemical reaction network (CRN). We will conclude our paper by showing how a very simple  chemical system can be trained and used to perform multi-label classification. . 
\section{Methods}
\subsection{The generalised neuron model}
The generalised neuron model (GNM) is a parametrised family of models that can be tuned to display varying degrees of spikiness, temporal autocorrelation of the membrane potential, and hysteresis (see fig. \ref{detailpics2}). The model is defined by the update function of the membrane potential  $V(t)$. If updates are made in discrete time, as is usual in the SNN literature, then the model is as follows: 
\begin{subequations}
\begin{align}
V(t) - V(t-1) &= I(t) - \underbrace{(\eta \gamma R(t-1) V(t-1) + (1-\eta) \alpha V(t-1))}_{:=D(t)}
\nonumber
\\
R(t) - R(t-1)&= \zeta \frac{V(t-1)^{h}}{\vartheta^{h}_{\textrm{B}} + V(t-1)^{h}} - \beta R(t-1) \label{VR_cont}
\end{align}
Here, $I(t):=\displaystyle\sum_{i=1}^M w_i \delta(t^i_j -t)$ is the sum of weighted inputs at time $t$ and $t^i_j$ is the time of the $j$-th spike of input $i$, where $i$ runs from 1 to $M$; $\delta(x)$ is 1 if $x=0$ and 0 otherwise; $0\leq w_i\leq 1$ is the weight of the $i$-th input. $\alpha$ and $\gamma$ are decay coefficients of the membrane potential, $\vartheta_{\textrm{B}}$ is a behavioural threshold, and $0 \leq \eta\leq 1$ is a model choice parameter, $\zeta$ is a rate parameter of the Hill function, $h$ is a Hill function coefficient, and $\beta$ defines the decay rate of $R$. The model is best understood by considering some special parameter choices. 
\par
For $\eta=0$ the update function of the membrane potential reduces to a leaky integrator:
\begin{equation}
 V(t) - V(t-1) = I(t) - \alpha V(t-1) \label{minmodel}
\end{equation}
In this case, the membrane potential is increased by whatever the input $I$ is at time $t$, and it decays by a constant factor $\alpha$. In the limiting case of $\alpha=0$ there is no decay at all.  The opposite extreme  is $\alpha=1$ which means that at each time step all the previous membrane potential is forgotten. For the discrete model it is not meaningful to set $\alpha > 1$ or $\alpha < 0$. The parameter $\alpha$ determines the temporal autocorrelation of the membrane potential, or the ``memory''. We will find this to be a crucial parameter for the performance of the GNM. 
\par
When $\eta>0$, then an additional, time dependent decay rate $R$ becomes relevant. $R$ can be thought of as describing the number of ion-channels that only open after the membrane potential approaches the  threshold $\vartheta_\textrm{B}$ and  close, stochastically, with a rate $\beta$. Alternatively,  this can be understood as the simplest model that implements a soft post-spike reset.  

\begin{figure}
\centering
\subfloat[hysteresis1][]{ \includegraphics[width=0.47\textwidth]{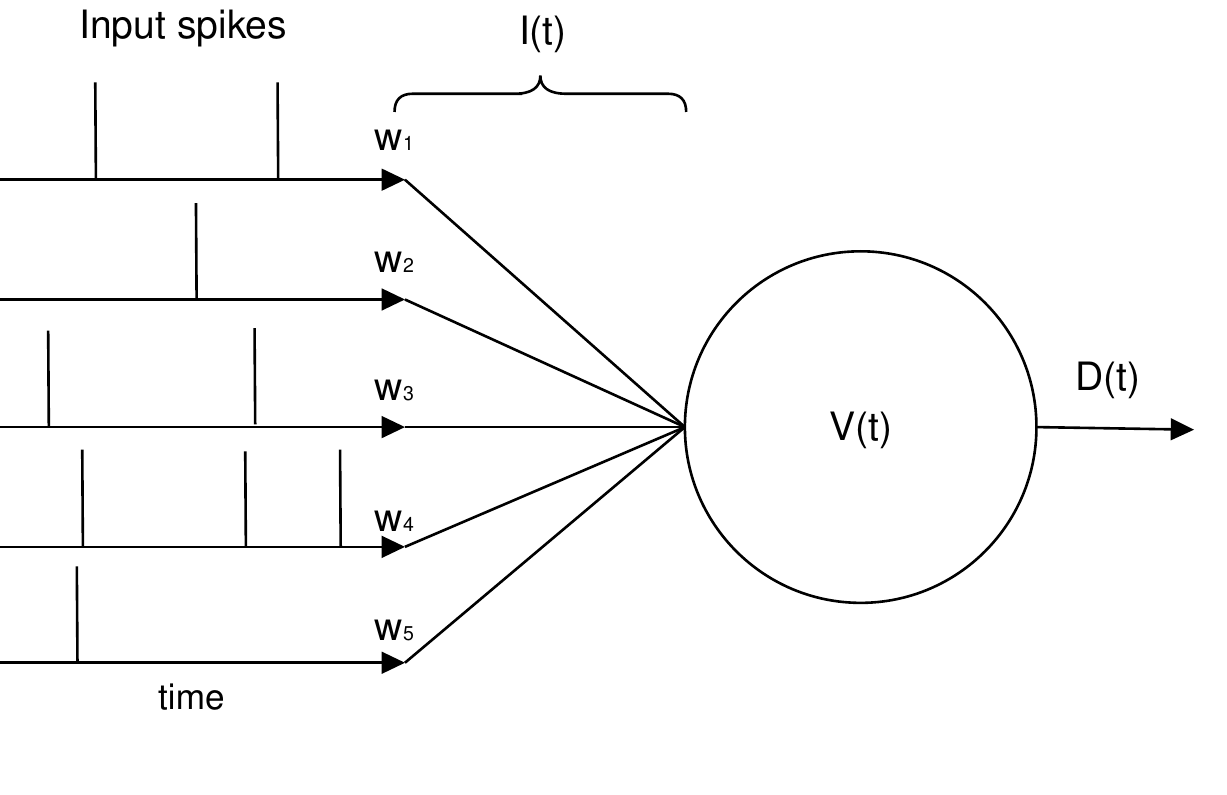} \label{neur_ex1}}
\subfloat[hysteresis2][]{ \includegraphics[width=0.47\textwidth]{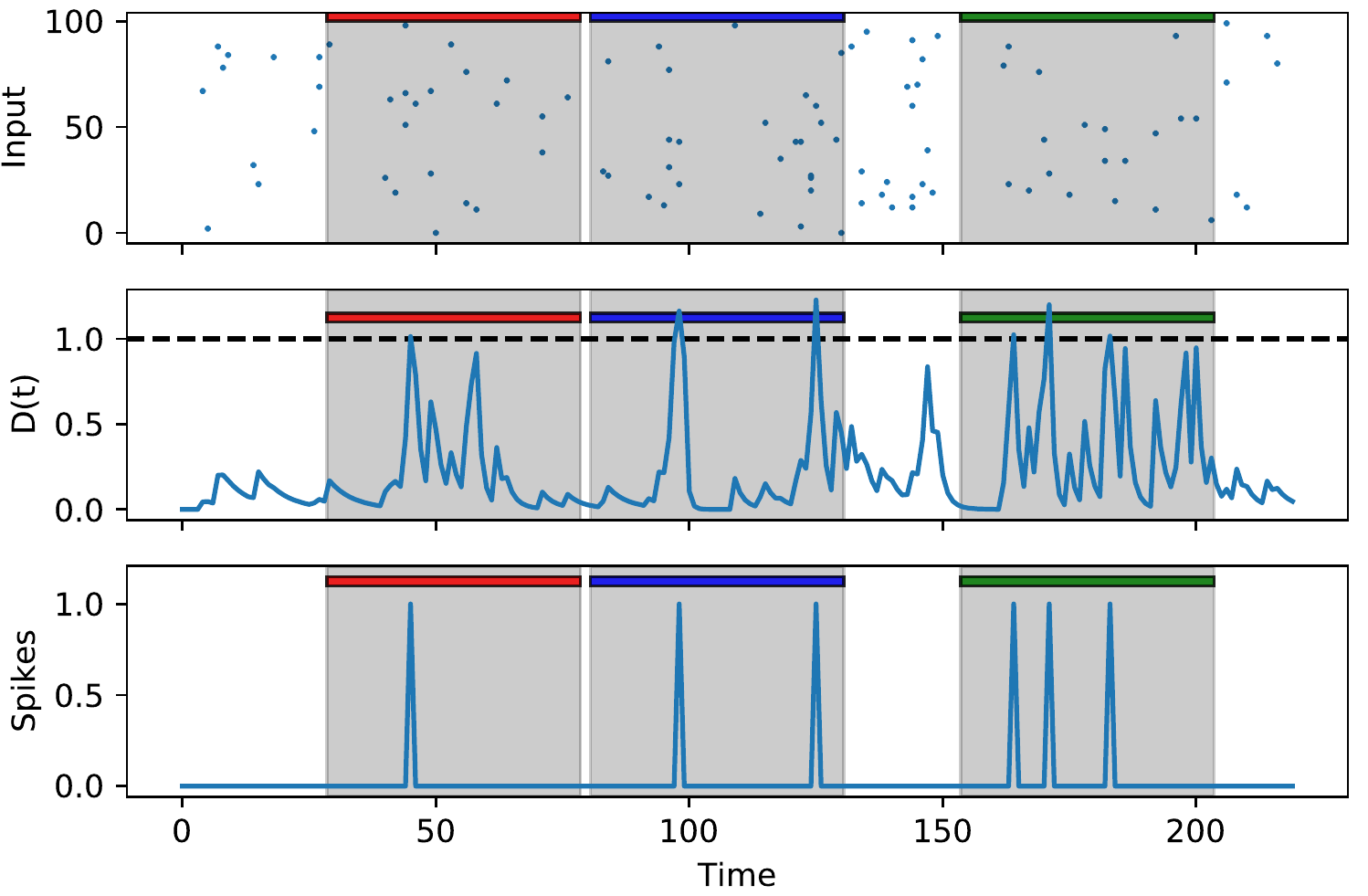} \label{neur_ex2}}\\
\subfloat[hysteresis3][]{ \includegraphics[width=0.47\textwidth]{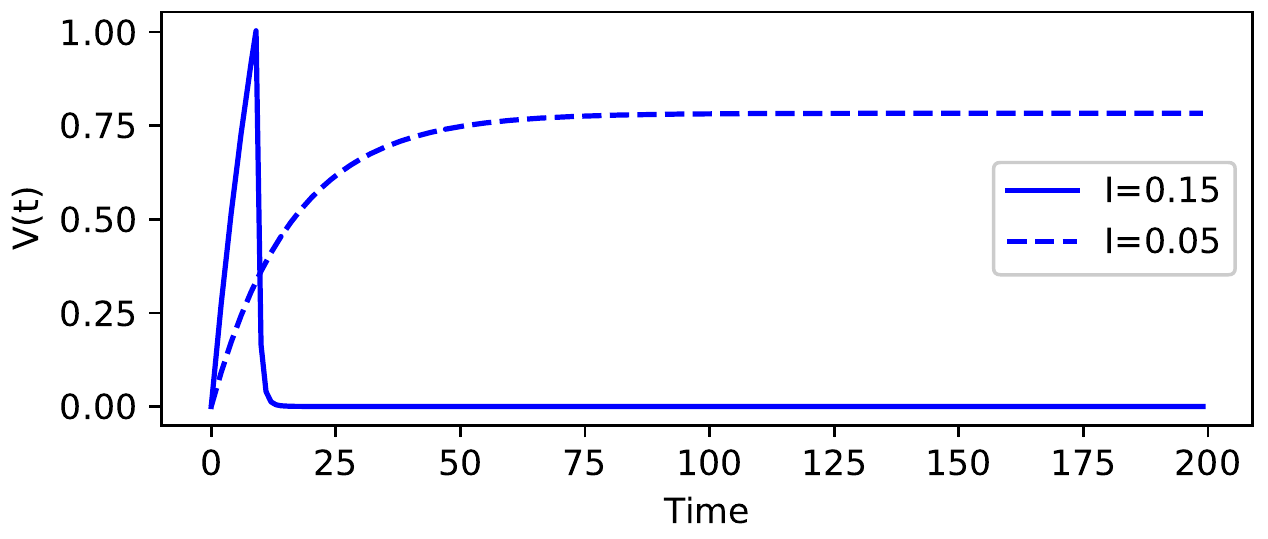} \label{hysteresis1}}
\subfloat[hysteresis4][]{ \includegraphics[width=0.47\textwidth]{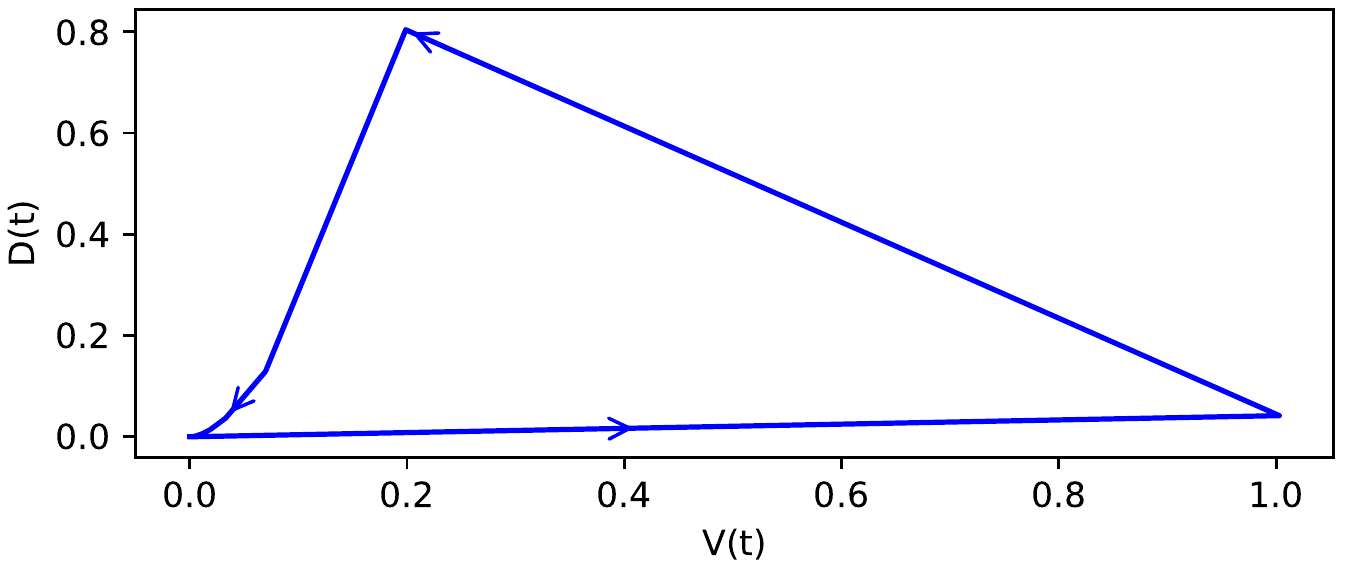} \label{hysteresis2}}
\caption{\protect\subref{neur_ex1} Schematic outline of the GNM model. GNM has $N$ weighted input connections, each of which receives temporal patterns of spikes. \protect\subref{neur_ex2} When the readout function reaches a threshold  $\vartheta_{\text{R}}$,  an output spike is recorded. In the example above, three patterns are presented, red, blue and green. The GNM responds with a single spike to the red pattern, two spikes to the blue pattern, and three to the green pattern. It should remain silent otherwise during a noisy phase.  \protect\subref{hysteresis1} Membrane potential as a function of time of the GNM stimulated by sub-threshold (dashed line) and super-threshold (solid line) continuous input, with parameters $\eta=0.8$, $\alpha=0.3$, $\beta=0.1$. \protect\subref{hysteresis2} Hysteretic behaviour of $D(t)$ as a function of $V(t)$ in the presence of super-threshold input. 
}
\label{detailpics2}
\end{figure}

\par
At the start of a simulation, $R(0)$ will be set to 0. The subsequent increase of  $R(t)$ depends on the membrane potential via a Hill-function (first term on the right hand side of eq.~\ref{VR_cont}), which is a sigmoidal activation function.  As $h\to\infty$ the Hill function approaches a step-function  with a transition at the  point $V=\vartheta_\textrm{B}$. Even for finite values of $h$ the Hill function will be close to zero (one) when the membrane potential $V(t)$ is  below (above)  $\vartheta_\textrm{B}$. Note that the decay rate $R$  decays itself with a rate of $\beta$. 
\par
The effect of this additional decay mode is that the membrane potential may decay faster after having crossed a threshold value $\vartheta_\textrm{B}$. This introduces a ``memory'' about past spike events into the model. The duration of the reset depends on the value of $\beta$, and continues even if the membrane potential falls back below the threshold. Thus, the model has {\em hysteresis}; see fig. \ref{hysteresis2}. 
\par
Before continuing it is useful to discuss briefly the relation between the GNM model and other well known neuronal models: Unlike the MST, the update rule of the  GNM is  purely state dependent. The $\eta$ parameter can be seen as regulating the ``spikiness'' of the model. The soft reset of the MST can be simulated in the GNM when $\eta>0$ and the values of the parameters $\beta, \zeta$ are set appropriately. The well known LIF neuron  behaves like  the GNM model with $h=\infty, \zeta=1, \gamma=1/\eta$ and $0<\eta<1$ up to reaching the  behavioural threshold $\vartheta_\textrm{B}$, including the post-spike reset. However, following the reset, the LIF undergoes a deterministic refractory period, during which it remains insensitive to inputs. This type of fixed refractory period cannot be simulated by the GNM. 
\par
In addition to the discrete time dynamics of eq. \ref{VR_cont}, below we will also show simulations of the full continuous time dynamics. In order to describe the continuous dynamics, we need to convert the difference equations eq. \ref{VR_cont} into proper differential equations, thus obtaining:
\begin{align}
\frac{d}{dt} V(t) &= I(t) - \underbrace{(\eta \gamma R(t) V(t) + (1-\eta) \alpha V(t))}_{:=D(t)}
\nonumber
\\
\frac{d}{dt} R(t) &=  \zeta \frac{V(t)^{h}}{\vartheta^{h}_{\textrm{B}} + V(t)^{h}}  - \beta R_{i}(t)
\label{contmodel}
\end{align}
In the differential equation model, the parameters $\alpha, \beta, \gamma, \zeta$  become rates and are restricted to be positive, although they may be greater than 1. The model choice parameter, however, remains restricted to $0 \leq\eta\leq 1$.
 See fig. \ref{neur_ex1} for a graphical representation of the model.
\par
In the case of $\eta=0$ (i.e. no spikiness) the full model (eq. \ref{contmodel}) reduces to:
\begin{equation}
\dot V= I - \alpha V
\label{minmodelc}
\end{equation}
\end{subequations}
\subsection{Quantifying ``spikes''}
\label{agg_label_error}

In the discrete time version of the GNM ``spikes'' are determined by counting how often the membrane potential $V(t)$ (or decay $D(t)$) crosses the readout threshold $\vartheta_{\text{R}}$ from below. In multi-label classification, this value indicates class membership. In the case of a single pattern, we require it to cross the threshold exactly once. 
The error is then simply the difference between a target number of output spikes and the actual number of spikes during $M$ timebins of a pattern.
\par
\begin{figure}
\centering
\includegraphics[width=0.9\textwidth]{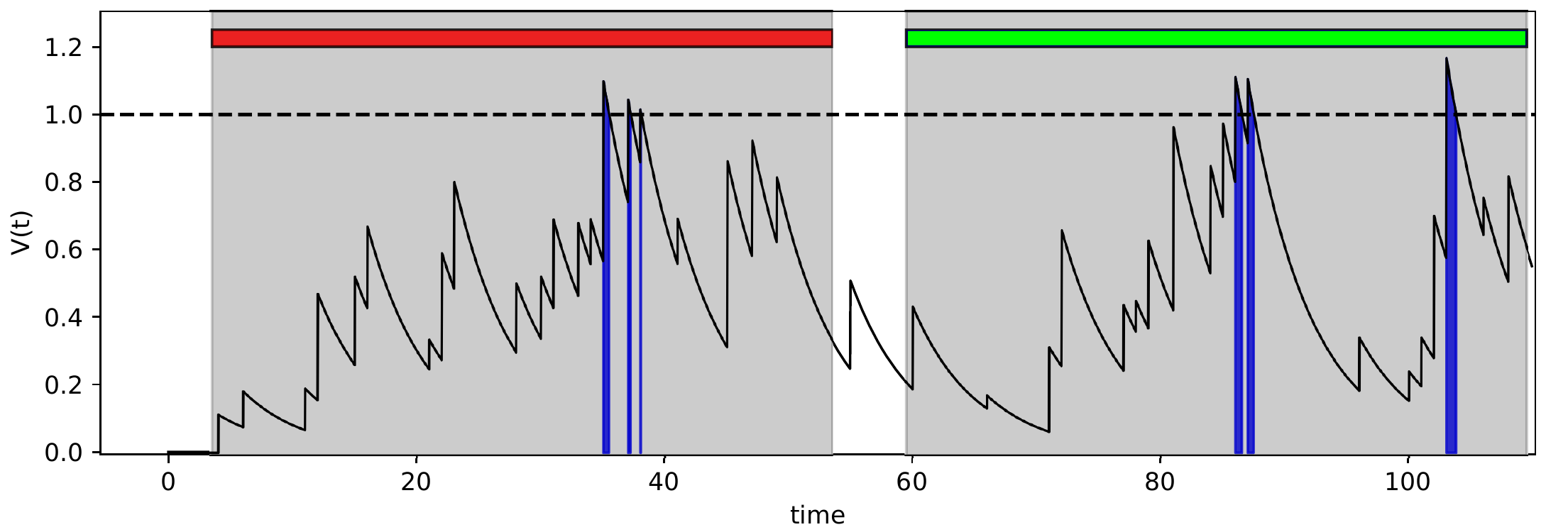}
\caption{ 
Example of pattern representations learnt in a continuous-time GNM trained in aggregate label setup. The integral of super-threshold membrane potential $\mathcal{S}$ is marked in blue. In this case the continuous spike integral totals at $\sim0.81$ for the red pattern, and $\sim1.98$ for the green pattern. In order to determine the spike indicator, we integrate the membrane potential when it is above a threshold. The total integral then indicates the number of spikes. In the case of the red pattern, the membrane potential crosses the threshold 3 times, but the total integral is $\approx 1$, hence it indicates a single spike (rather than 3). }
\label{fig:cont_example}
\end{figure}

In the continuous time case we need to use a different way to quantify spiking based on the integral of the GNM membrane potential when it exceeds $\vartheta_{\textrm{R}}$. 
\begin{equation}
     \mathcal{S} := \int_{0}^{T} \Theta (V(t) - \vartheta_{\textrm{R}})  V(t) \: dt
    \label{Integral_outputs_D}
\end{equation}
where $\Theta$ denotes the Heaviside function, $\vartheta_{\textrm{R}}$ is a readout threshold, and $T$ is the length of the trial.  The error is then defined as the difference between the actual and the desired spike output; see fig. \ref{fig:cont_example}.
\subsection{Training algorithms}

We will use three different training algorithms to compare the GNM with the MST. Firstly, we will use a version of the eligibility-based ALL algorithm proposed by G\"utig, adapted to the GNM. We will then also show that there are alternative algorithms which can be applied to the GNM and provide better performance.

\paragraph{Aggregate-label learning (ALL)}
\label{methods:all}
We first describe the eligibility-based learning algorithm similar to the one proposed by \cite{Gutig2016}. We update the weights after a trial  during which the neuron is shown  target patterns and noise  for a given period of time. Depending on what was shown, a target spiking number is then determined.  After the trial is finished, we then set the error to a negative value if the neuron spiked too many times during this trial, to a positive value if it didn't spike enough, and otherwise we don't update. Note, that the algorithm does not provide any feedback about the degree to which the output was wrong, only about the ``sign'' of the error. The learning step proceeds by updating the weights like so:
\begin{align}
    \label{eq:MultiTempCorrelationDeltaW}
    \Delta \omega_{i} &= \left\{\begin{matrix} \pm\lambda  &\textup{if} \, \varepsilon_{i} > D_{9} \\
    0 &  \textup{if} \, \varepsilon_{i} \leq D_{9}
    \end{matrix}\right.
     \nonumber
    \\
      \varepsilon_{i} &:= \int_{0}^{T} I_{i}(t) V(t) \: dt 
\end{align}
Here, $\lambda$ denotes the learning rate which is positive when the error is positive and negative otherwise, $D_{9}$ represents the 9th decile (top 10\%) of the most eligible synapses. The variable $\varepsilon_i$ is the eligibility of a pre-synaptic neuron $i$ towards the  post-synaptic neuron. It quantifies the extent to which the pre-synaptic neuron $i$ has contributed to a spike of the post-synaptic neuron. 
\paragraph{Error trace learning (ET)}
\label{one_error_trace}
Secondly, we introduce an additional new training approach which utilises precise information about the timing of erroneous spikes, as well as when and which feature patterns should have been recognised.
The way the weight updates are calculated are the same as in the ALL algorithm, except that now we based the eligibility and the error on the values of the integral at the time when each of the patterns was presented, rather than after an entire episode of patterns and noise sequences. Thus, the algorithm obtains more detailed information about which weights caused erroneous spiking.
This means that in response to a pattern with a target of two spikes, the neuron is supposed to cross the readout threshold  exactly twice during the duration of the presented pattern.  
We calculate the error for each individual synapse by correlating its inputs with the error trace:
\begin{align}
    \label{eq:MultiTempCorrelationDeltaW}
    \Delta \omega_{i} &=   \lambda \varepsilon_{i}
     \nonumber
    \\
      \varepsilon_{i} &:= \int_{0}^{T} I_{i}(t) E(t) \: dt 
\end{align}
where $\lambda$ is again the learning rate, and $E(t)$ denotes the error trace. Here, the variable $\varepsilon_i$ should be understood as ``error blame'' of a pre-synaptic neuron $i$ towards the post-synaptic neuron. It quantifies the extent to which the pre-synaptic neuron $i$ has contributed to the erroneous activity of the post-synaptic neuron.

\begin{figure}
\centering
\includegraphics[width=0.7\textwidth]{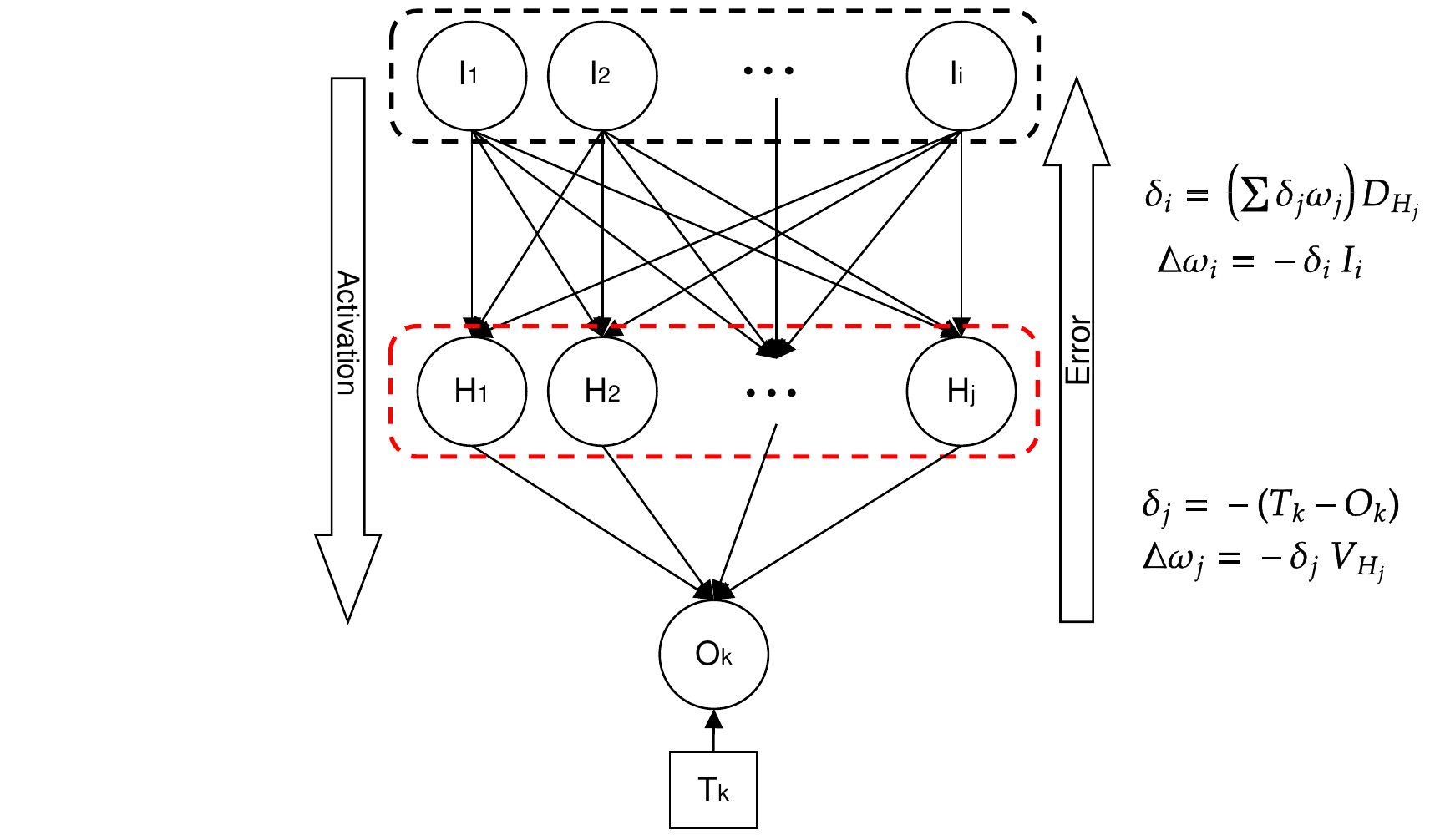}
\caption{Error backpropagation in multi-layered network of the GNM. 
We introduce an intermediate layer of 10 hidden neurons. The hidden neurons have the same neural dynamics as the output neuron, and connect to the input source in all-to-all fashion. It is worth noting that the output of the hidden neurons is presented to the output neuron as a sum of decaying currents in the continuous form.  In order to successfully solve this task, we need to be able to propagate the error back through the network, and correctly adjust the weights for both the output neuron and the hidden layer.  
The hidden layer neurons in red dashed box are subject to continuous lateral inhibition. 
}
\label{fig:BPa}
\end{figure}
\paragraph{Error trace backpropagation learning (BP)}
\label{backprop}
Finally, unlike traditional spiking models, where backpropagation can only be applied indirectly \citep{Neftci2019, DBPL}, the activation function of the GNM is differentiable and backpropagation can in principle be applied with no constraints. 
The temporal precision of error signalling in ET enables us to further extend it to training multi-layered networks of GNMs. 
By way of demonstrating this, we show the network's ability to solve the multi-label classification task using an architecture of layered GNMs consisting of 10 hidden neurons (see fig. \ref{fig:BPa} for the architecture details). 
\subsection{Momentum heuristic}

In order to improve the speed of learning, we also use a \emph{momentum heuristic}. 
During each learning step we add a fraction of previous synaptic change to the update value:

\begin{equation} \label{eq:MultiTempCorrelationDeltaW}
    \Delta \omega_{i}^{\textrm{current}} = \omega_{i} + \gamma \Delta \omega_{i}^{\textrm{previous}}
\end{equation} 

where $\gamma$ is the momentum parameter; in all  experiments it is set to 0.2.
\section{Results}
\subsection{Aggregate label learning in the GNM}

We first tested how well the discrete GNM can learn a single spatio-temporal pattern. Such a pattern is a temporal sequence of $M$ binary strings of length $N$. Throughout this article we kept $N=100$ and $M=50$. Patterns were generated randomly by drawing each of the bits from a Bernoulli distribution with $p(1)=0.005$. In addition to randomly generated but fixed patterns, we expose the neuron to a stream of noisy background activity. The random activity is generated in the same way as the patterns, but unlike it, the noise is produced at each timebin. As a consequence, the statical properties of noise and pattern are identical in this setup. 
\par
The first task we set is as follows: GNM should respond with exactly one spike if the input is a pattern and should stay inactive otherwise, i.e. if presented with noise. 
Unlike the MST, the GNM does not have discrete output spikes.
In order to interpret the output of the GNM we thus need to set an (arbitrary) readout threshold value  $\vartheta_{\text{R}}$.  The response of the GNM is determined by the number of times membrane potential $V(t)$ (or decay $D(t)$) crosses $\vartheta_{\text{R}}$ from below within the duration of the pattern; see fig. \ref{neur_ex2}. This number is used to indicate class membership. In the case of a single pattern, we require it to cross the threshold exactly once. 
\par
We train the neuron using the ALL algorithm (see section \ref{methods:all}). We first test the performance of the GNM on the task of learning a single pattern. Here, the neuron is presented with a random number of spatio-temporal patterns embedded into noisy background activity at random times. At the end of a trial the GNM receives feedback indicating whether it has released too many or too few spikes.
In all experiments we use the following parameters: $\beta=0.3$, $\zeta=1$, $\gamma=1$, which allow the neuron to exhibit a post-spike reset closely resembling that of the MST. 
For each target pattern we sampled 41 different values of both $\alpha$ and $\eta$ parameters (altogether 1681 parameter combinations), in order to test the performance of the GNM.
We varied systematically the model choice parameter $\eta$ from 0 (``no spikiness'') to 1 (``complete spikiness'') and the decay rate  $\alpha$ from 0 (``complete memory'') to 1 (``no memory''). 
For each combination of parameters, we trained the GNM over 60000 epochs (trials consisting of a random number of patterns embedded into randomly generated noisy background activity) with a learning rate $\lambda=0.0001$. 
\par
In order to determine the quality of learning, we subjected the trained GNM  to a stream of noise with randomly interspersed target patterns. If working correctly, the GNM should not respond to the noise, but should respond to the pattern. In practice, GNMs will not function perfectly. In order to quantify the classification reliability of the neuron, we recorded the number of random inputs given to the GNM before the GNM failed and averaged this number over 100 repetitions. We will henceforth refer to this as the {\em noisy performance measure} and use it as an indicator for the quality of the GNM solution. Here, a higher noisy performance is better.
\par
\begin{figure}[]
\centering{
\subfloat[fea1a][]{ \includegraphics[width=0.45\textwidth]{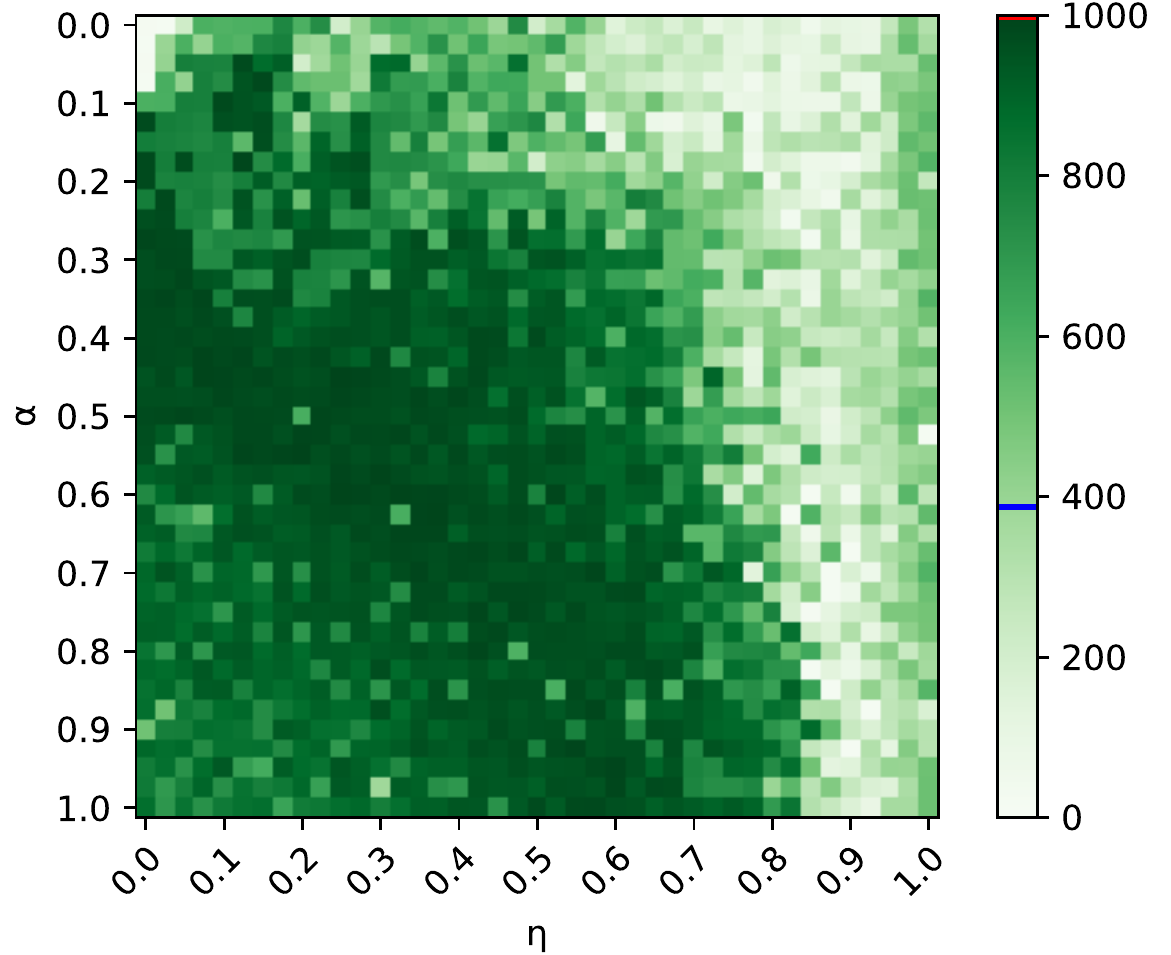}\label{fig:fea1a}}
\subfloat[fea1b][]{ \includegraphics[width=0.45\textwidth]{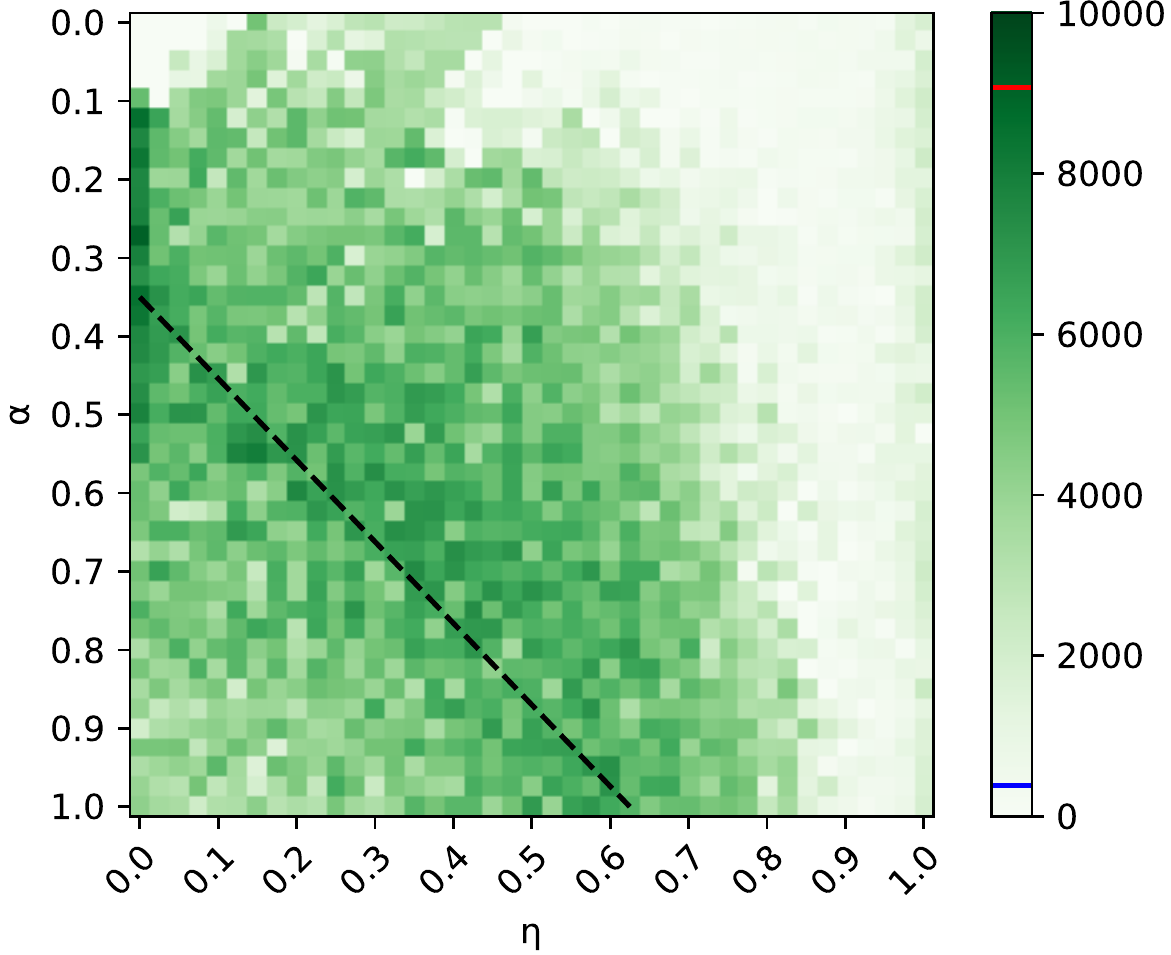}\label{fig:fea1b}}}
\centering
\subfloat[fea1c][]{ \includegraphics[width=0.9\textwidth]{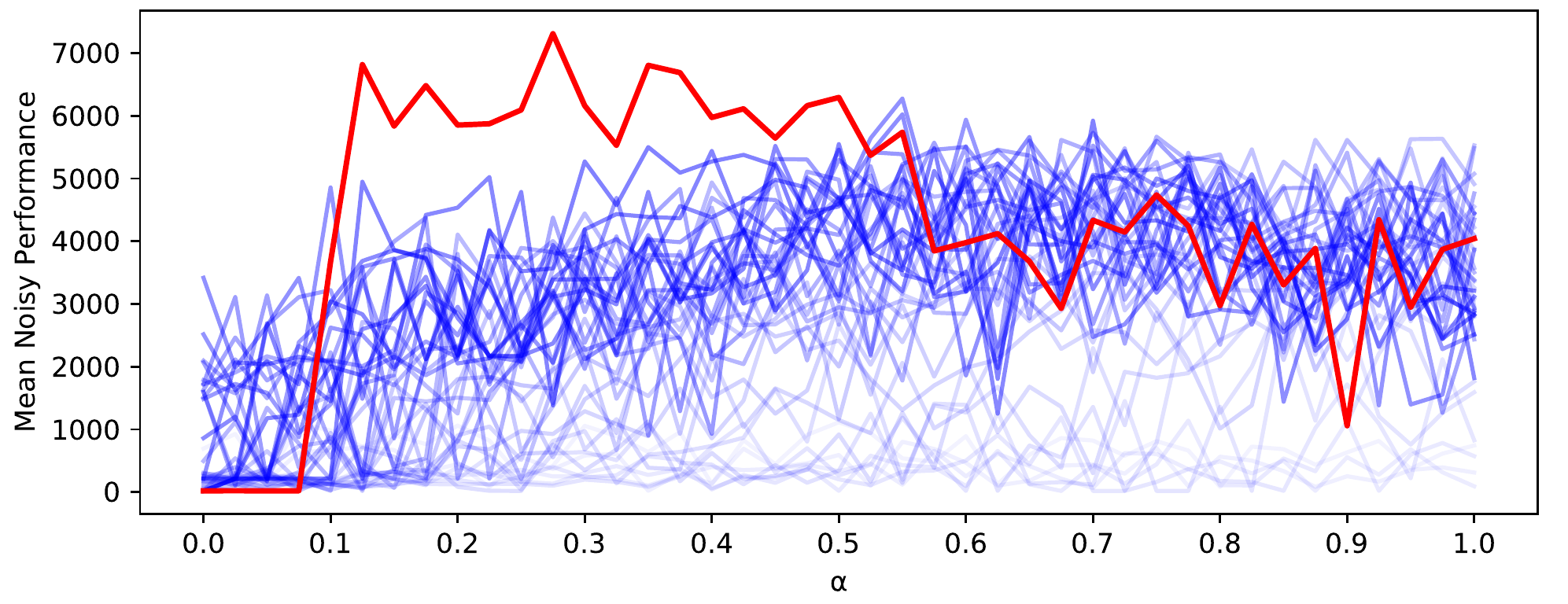}\label{fig:fea1c}}
\caption{Mean noisy performance averaged over 5 training iterations as a function of $\protect\alpha$ and \protect$\eta$.
The value reported is the average number of epochs a neuron can withstand without making an error, capped at \protect\subref{fig:fea1a} 1000, and \protect\subref{fig:fea1b} 10000 timebins. Each simulation was trained for 60000 epochs.
The corresponding MST performance averaged at $\sim377$ epochs, and is marked on the colorbar in blue. The red mark indicates the maximum value in the heatmap. The dashed line in \protect\subref{fig:fea1b} shows the estimated {\em optimal line} (see main text for explanation). \protect\subref{fig:fea1c} Noisy performance as a function of $\alpha$. The red line indicates $\eta=0$, and the blue lines show 40 other $\eta$ settings.}
\label{fig:oneexp}     
\end{figure}
\par
Fig. \ref{fig:oneexp} summarises the noisy performance of the GNM averaged over 5 different patterns representing altogether 8405 different training iterations of the GNM. The graphical representation reveals a qualitative structure of the parameter space, which we find to be generic for any number of patterns.
\par
For $\alpha, \eta \approx 0$, corresponding to the top left corner, noisy performance is low, i.e. the GNM does not classify well.  The reason for the  poor classification can be understood easily:  In this region, the decay of the membrane potential $V(t)$ is low and the GNM integrates over all past events. The membrane potential remains in a permanent super-threshold state and thus unable to cross the threshold $\vartheta_\textrm{R}$ from below (or indeed from above). 
\par
Allowing some leak by increasing $\alpha$ while keeping $\eta$ at 0 (i.e. going down the left-most column in fig. \ref{fig:oneexp})  improves the performance dramatically. For example, by adjusting  $\alpha$ from 0.05 to just 0.08 the noisy performance increases from approximately 0 to the global best. As  $\alpha$ approaches 1 the performance decreases again.
Therefore, for the sub-family of GNM models with $\eta=0$, there must be a value of $\alpha$ that optimises the learning, although this optimum is not well resolved.
Note that in the region $\eta=0$ that we considered so far,  the neural dynamics is reduced to $V_{i}(t+1) = V(t) + I - \alpha V_{i}(t)$.  As such, it lacks entirely the features that are usually associated with spiking neurons, including discrete spikes and an activation threshold.
\par
A behavioural threshold $\vartheta_\textrm{B}$  is introduced to the GNM by increasing the spikiness parameter $\eta$. Fig. \ref{fig:oneexp} reveals that high performance of the model concentrates along a fuzzy line of combinations of  $\alpha$ and $\eta$ --- we will henceforth refer to this as the {\em optimal line} (see fig.~\ref{fig:fea1b}). However, performance along this line does not significantly increase for $\eta > 0$ relative to the non-spiking case of $\eta=0$.
Indeed, we can see that the GNM performance drops for extreme values of $\eta\approx 1$. In this regime, the threshold dynamics dominates and the membrane potential is constrained to a small range of values, making learning impossible.
Based on this we conclude that, at least for the case of a single pattern, introducing spikiness does not bring any benefits.
Globally best performance can be achieved for $\eta=0$ at $\alpha \simeq 0.3$. 
\par
There is also an appealing conjecture for the origins of the optimal line:  We observe from  eq. \ref{VR_cont}  that the decay is effectively reduced by $(1-\eta)$. The optimal line can then be interpreted as a consequence of there being an optimal value for the parameter $\alpha$. To see this, assume that this optimal value is given by  $\alpha=\alpha^*$. Assume further, that the actual value of $\alpha$ is set to $\alpha'>\alpha^*$. A suitable choice of $\eta$ satisfying $(1-\eta)=\alpha^*/\alpha'$  can effectively offset the non-optimal choice of $\alpha$ back to the optimal value. If true, this would generate precisely the observed optimal line in the parameter space portrait. Beyond this correction of the decay parameter, an increased spikiness has no apparent benefit.

\subsection{Multi-pattern learning in the GNM}

So far, we have only tested how the GNM learns a single pattern. The key achievement of the MST is that a single neuron can learn to recognise multiple patterns and multiple classes of patterns. For example, there may be a set of patterns to which the MST responds with one spike, and a set of patterns to which it responds with two spikes and so on. We now test whether the GNM can do the same. Similar to before, we interpret the GNM as ``spiking'' $n$ times if during the presentation of the pattern the membrane potential crosses the readout threshold $\vartheta_{\text{R}}$ from below $n$ times.
\begin{figure}
\centering{
\subfloat[fea2][]{ \includegraphics[width=0.334\textwidth]{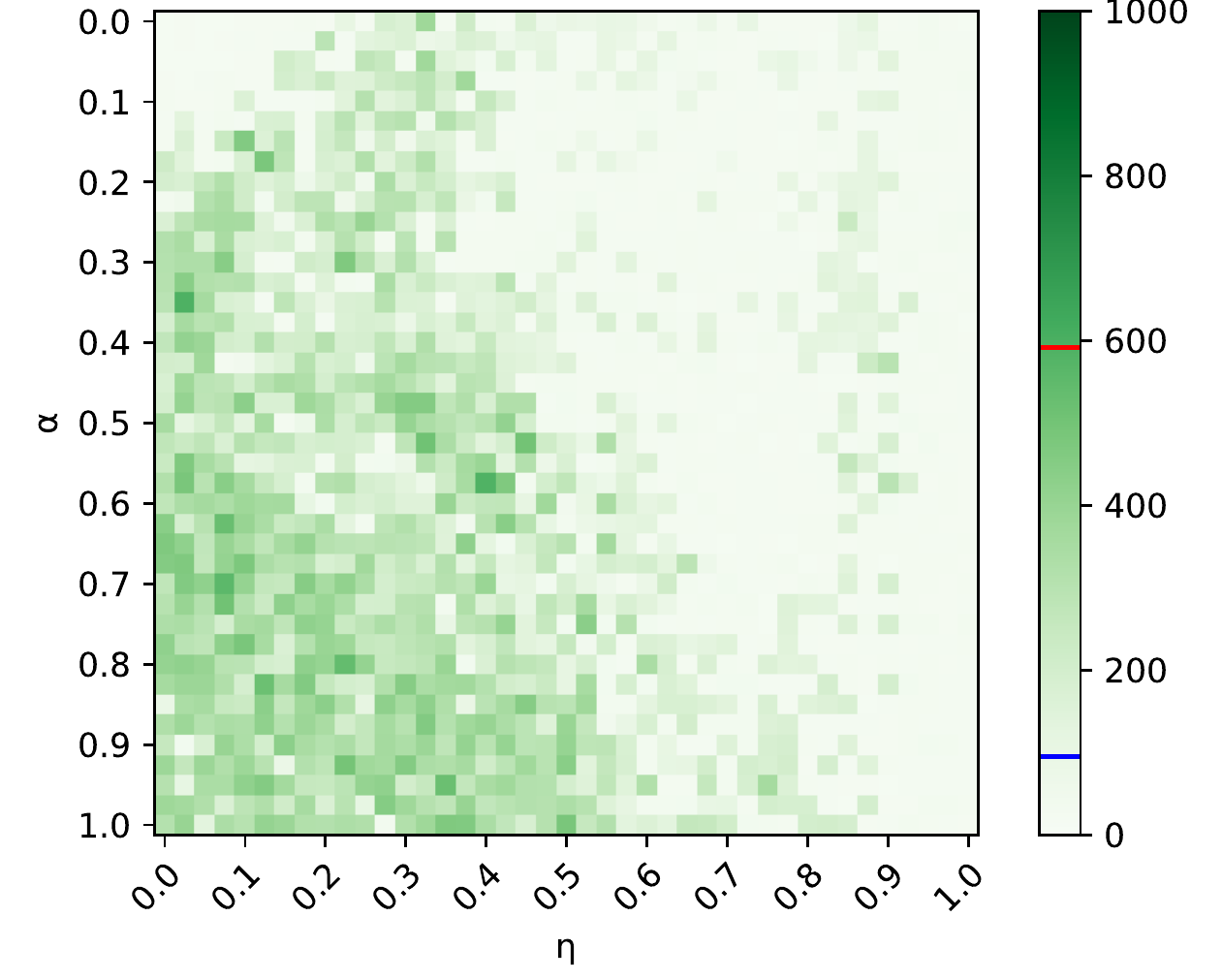}\label{fea2}}
\subfloat[fea3][]{ \includegraphics[width=0.332\textwidth]{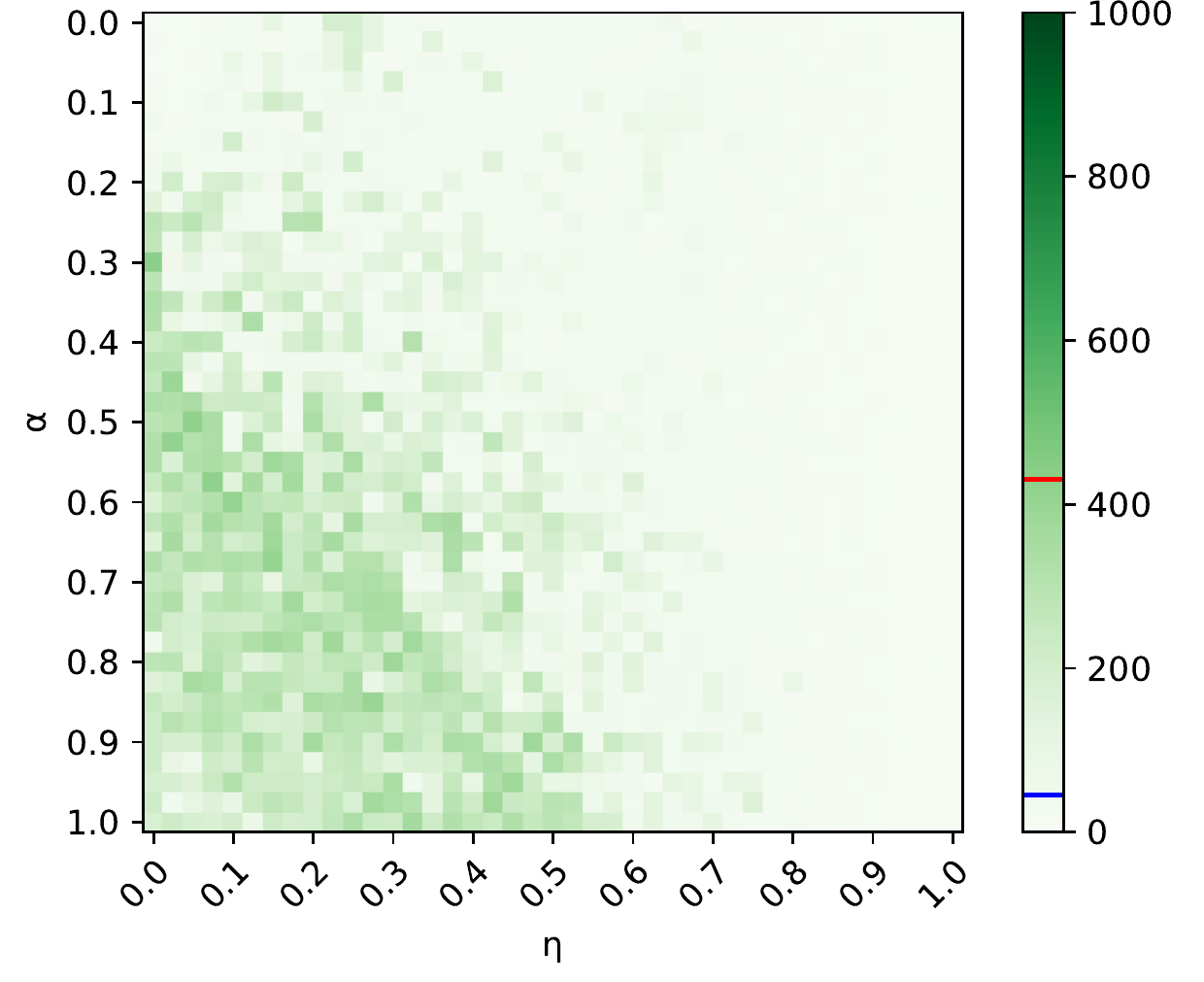}\label{fea3}}
\subfloat[fea4][]{ \includegraphics[width=0.332\textwidth]{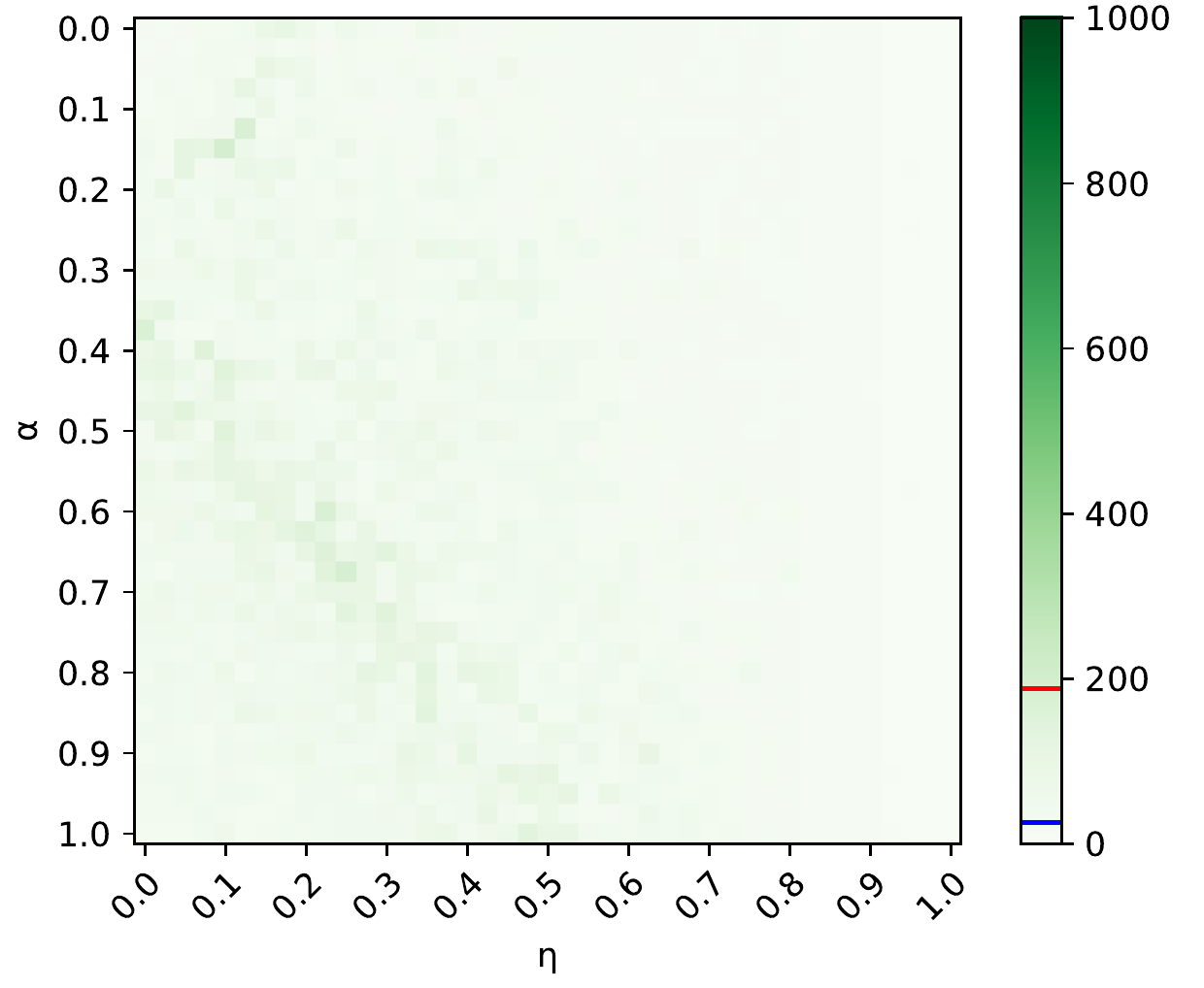} \label{fea4}}}
\caption{As fig. \ref{fig:oneexp}, but for \protect\subref{fea2}  two, \protect\subref{fea3} three, and \protect\subref{fea4} four classes of patterns. }
\label{manyexp}
\end{figure}
\begin{figure}
\centering
\subfloat[optimal_a][]{ \includegraphics[width=0.9\textwidth]{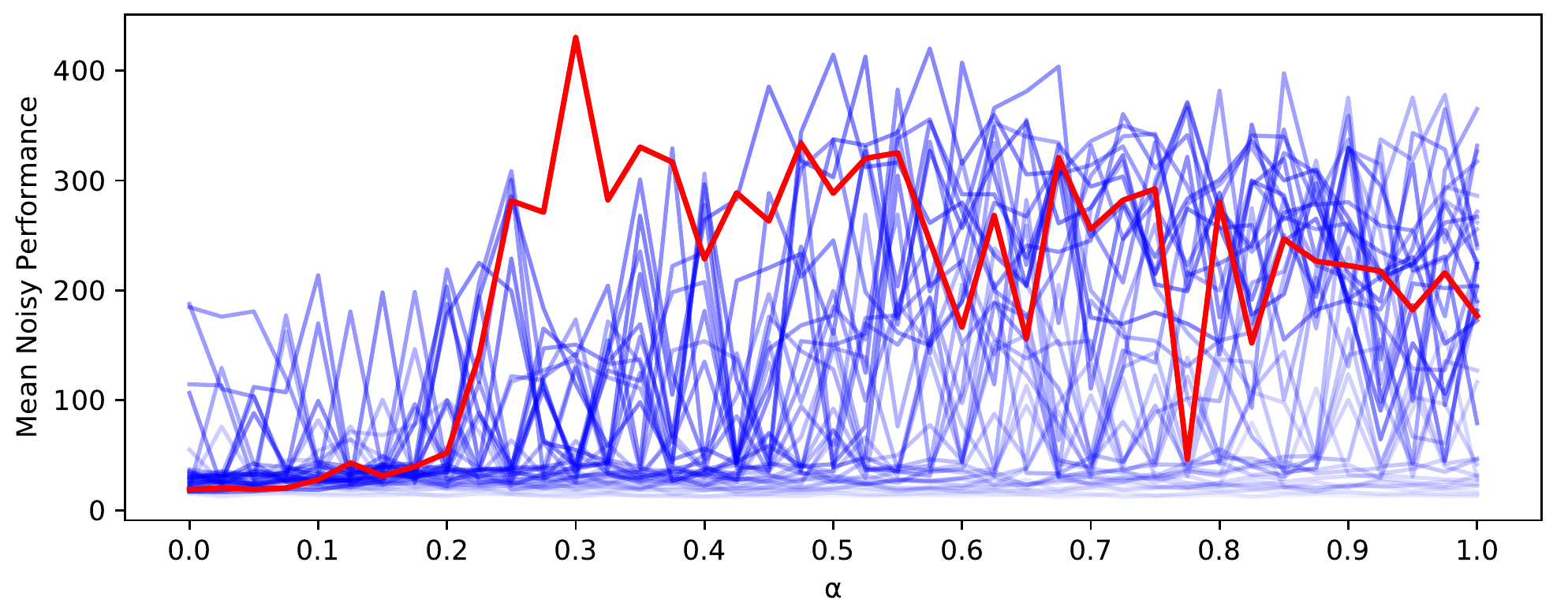} \label{optimal_a}}\\
\subfloat[residual][]{ \includegraphics[width=0.9\textwidth]{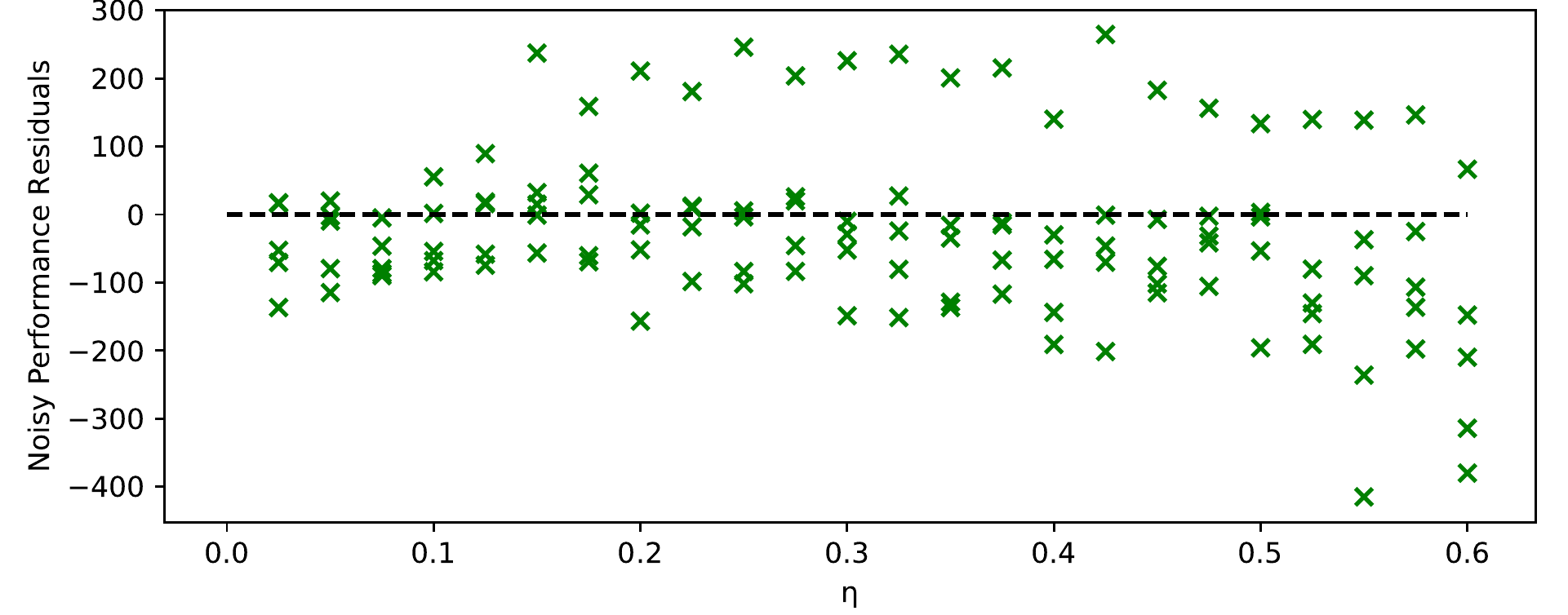} \label{residual}}
\caption{
\protect\subref{optimal_a} Noisy performance as a function of $\alpha$ for same data as in fig. \ref{manyexp}. The red line indicates $\eta=0$, and the blue lines show 40 other $\eta$ settings.
\protect\subref{residual} Noisy performance residuals for each of 5 training simulations as a function of $\eta$ in range from 0.025 to 0.6. Residuals as a performance difference of the best $\alpha$ for any given $\eta$ with the best performer from $\eta=0$.  The deviation from the dashed line indicates the difference in performance in comparison to $\eta=0$. Negative values indicate that the best training for a particular $\eta$ was worse than $\eta=0$ (see supplementary information S2 for the same graph for 1, 2 and 4 classes of patterns). We find that only in a single run $\eta=0$ was suboptimal, thus conclude that there is a variation of performance that depends on the random seed given for pattern generation. 
}
\label{3fea_details}
\end{figure}
\par
Using this convention, we found that the multi-pattern case shows a qualitatively similar picture in parameter space as the single pattern case (see fig.~\ref{manyexp}), including an optimal line. Altogether, however, the noisy  performance of the GNM dropped quickly with the number of pattern classes. For example, in the case of 4 different classes, the GNM responds to noise, and thus fails, after approximately 200 timebins on average, in the best case; see fig. \ref{fea4}. 
Again, as in the single class case, there does not appear to be any benefit in increasing $\eta$ above 0.
While for some patterns the performance of the GNM with $\eta > 0$  is better, there was no consistent best value of $\eta$ and the best model with $\eta=0$ was always comparable to the globally best result; see fig. \ref{fea3} and fig. \ref{3fea_details}. 
In order to understand this in more detail, we plot the noisy performance residuals graph (see fig.~\ref{residual} and supplementary information S2). 
Here, we define residuals as a performance difference of the best $\alpha$ for any given $\eta$ with the best performer from $\eta=0$. 
In simple terms, for each column in a heatmap, we select the best row and compare it to $\eta=0$. 
Thus negative values of the y-axis, i.e. points below the dashed line,  indicate that the best performer for a particular $\eta$ was worse than that of $\eta=0$, for a given set of feature patterns.  We found that most of the points fall under the dashed line, indicating that $\eta=0$ is competitive. 
One of the training simulations has exceeded the line. Given that the setup is identical for all simulations, apart from the random seed and the pattern, we take this as an indication that there is no fundamental performance difference between the parameters with $\eta=0$ and those with a positive $\eta$.  In summary, we found that the GNM can perform well on the multi-label classification task with performance comparable to the more complicated MST model (indicated by the blue line on the colorbars in fig. \ref{fig:oneexp} and \ref{manyexp}). Intriguingly, we also found that it is sufficient to consider the simplest ``non-spiking'' case of the GNM corresponding to $\eta=0$. 
\par
\subsection{Comparison to other training methods and neural models}
\label{comparison}

Above we found that the GNM is competitive with the MST. However the comparison was unfair because we compared a large number of simulations to just a single parameter setting of the MST. We now investigate the performance of the GNM relative to other models with more rigour. 

\begin{figure}
\centering
\includegraphics[width=0.9\textwidth]{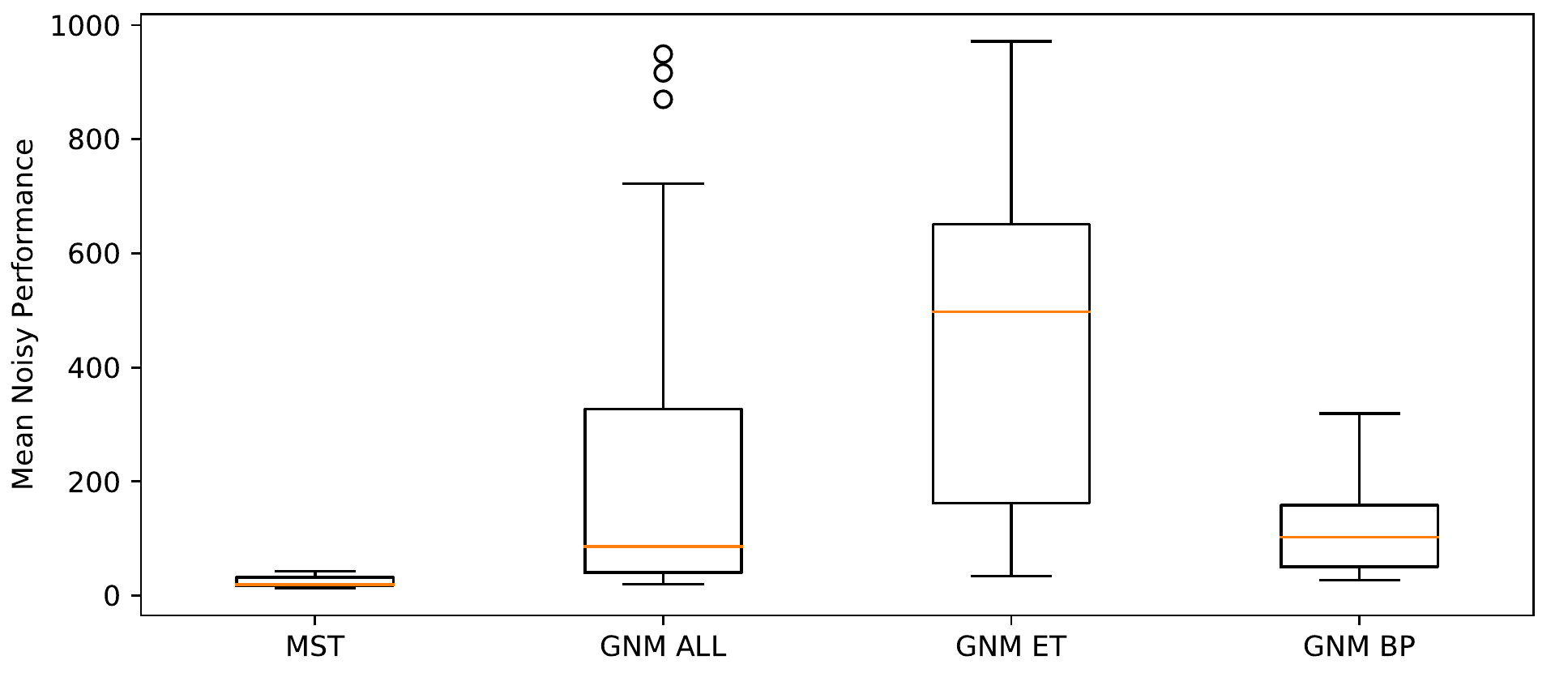} 
\caption{Comparison of the noisy performance measure in task involving classification of two classes of patterns for MST, GNM trained using ALL, GNM trained using ET, and multi-layered network of GNMs trained using BP. Neuron models have been trained for 60000 epochs with the following parameters: GNM: $\alpha=0.3$, $\eta=0$, $\beta=0.3$; MST: $\tau_{m}=20$, and $\tau_{s}=5$.}
\label{fig:comp}
\end{figure}

To do this we conducted 50 training simulations for the task of recognising two patterns and discovered that MST outperformed the GNM (trained using ALL) only 3 times, for fixed parameters of the GNM: $\eta=0.0$ and $\alpha=0.3$ and MST: $\tau_{m}=20$, and $\tau_{s}=5$ (see fig. \ref{fig:comp}).
This allows us to suggest that the GNM is not only simpler to implement but also learns better. 
\par
Next we performed the same comparison with the LIF neuron. The LIF neuron is similar to the GNM and it is reasonable to conjecture that it can be trained to perform multi-label classification as well. The MST and the GNM  differ from the LIF in two features: \one They do not have a refractory period and \two their reset function following a spike is exponential, rather than an absolute reset to the resting potential.  
\begin{figure}
\centering
\includegraphics[width=0.9\textwidth]{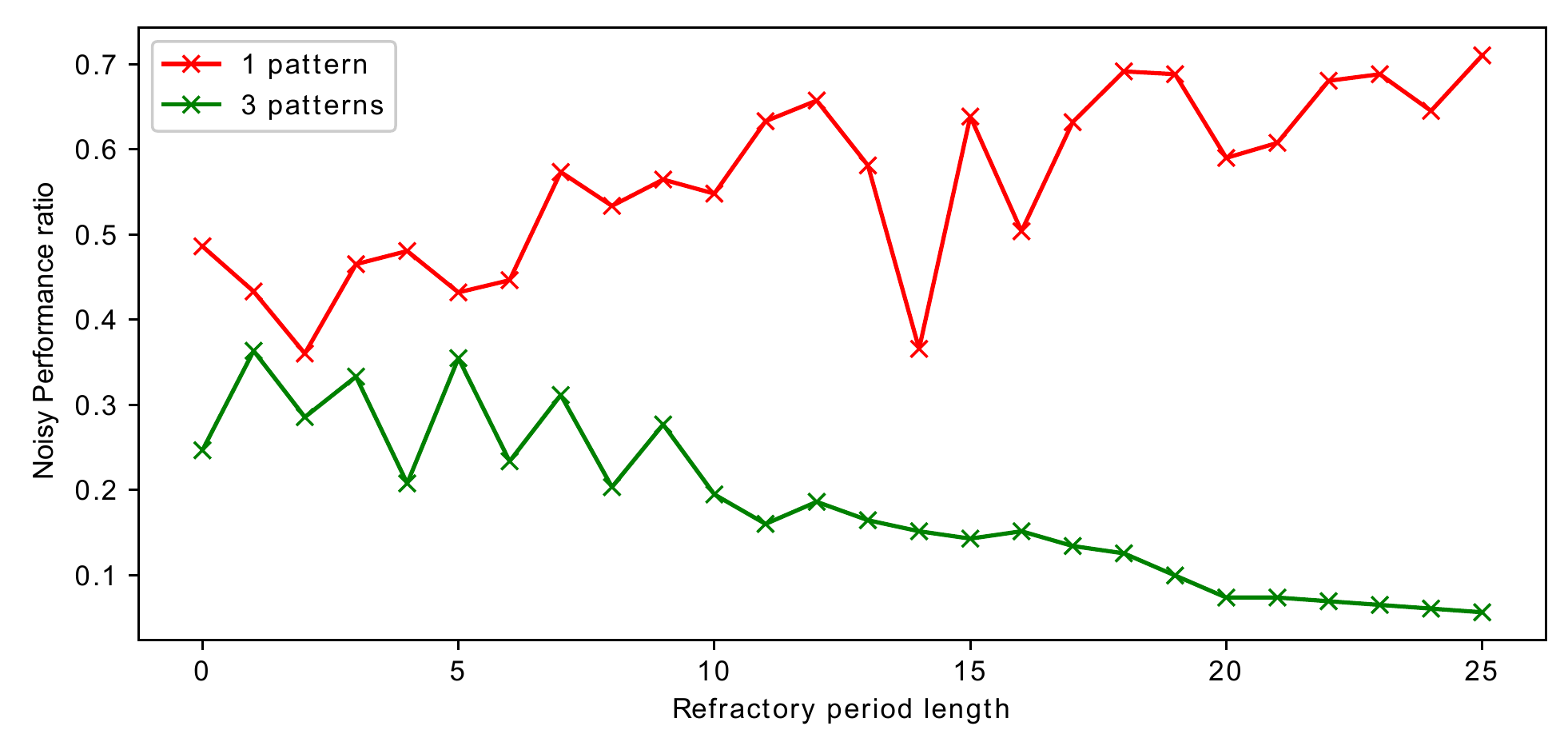} 
\caption{Noisy performance ratio of the GNM with $\eta=0$ with a corresponding LIF neuron on the task of learning one and three classes of patterns. The noisy performance ratio is defined as the average noisy performance of LIF neuron divided by that of the GNM (ALL). Both neuron models have been trained for 60000 epochs with the following parameters: GNM: $\alpha=0.3$, $\eta=0$, $\beta=0.3$; LIF: $\alpha=0.3$, and varied length of the refractory period.}
\label{fig:LiF_GNM}
\end{figure}
We compared the ability of the LIF neuron to recognise patterns with the GNM assuming $\eta=0$ and $\alpha=0.3$; see fig. \ref{fig:LiF_GNM}. We used the same parameters for the LIF neuron and varied the length of the refractory period from 0 to 25. We contrast the two neural models by calculating the noisy performance ratio, which is the average noisy performance of LIF neuron divided by that of the GNM in the same task. For a single pattern, the LIF neuron performs comparably to or slightly worse than the MST and the GNM, and the performance tends to increase with the length of refractory period (see fig. ~\ref{fig:LiF_GNM}). However, for more than one pattern, the LIF neuron normally yielded a worse noisy performance than both MST and GNM. In this case the performance drops with the increase in refractoriness. This shows that the period of forced inactivity hinders the multi-spike response. Note that for a refractory period of 0 the LIF neuron is identical to the GNM with the exception of the hard reset following a spike, but still performs worse on multi-label classifications. This suggests that the hard reset hinders multi-label classification and is the reason for reduced performance of the LIF model. This is consistent with our conjecture that the temporal autocorrelation of membrane potential is important for learning. 
We see from fig. \ref{fig:LiF_GNM} that the noisy performance increases with the refractory period. This is a consequence of the fact that during that period the LIF neuron remains insensitive to inputs. 
%
%
\par
In addition to comparing the GNM to other neural models trained using the same algorithm, we also contrasted the performance of the original ALL algorithm with training techniques involving more information about the time of the error, i.e. ``error trace learning'' (see section \ref{one_error_trace}). We find that the error trace learning algorithm has consistently outperformed ALL (see fig. \ref{fig:comp}). Out of 50 training simulations that we conducted (for a GNM neuron with parameters $\eta=0$, $\alpha=0.3$) 46 achieved a better result in terms of noisy performance when using ET method. This tells us that despite the fact that the ALL is an elegant and simple training rule, it is also suboptimal. 
\par
Moreover, we propose an extension of the ET rule to multi-layer networks setups. The error backpropagation algorithm (see section \ref{backprop}) can be applied directly to the GNM (see fig. \ref{fig:comp}). However, for the present task, we could not find any benefits in applying backpropagation. This is not to say that for more complex problems, backpropagation may be beneficial in SNNs. Exploring this is beyond the scope of this paper and we leave it to future research.
\par
\subsection{Interpreting GNM as a system of chemical reactions}
\par
A common assumption in the SNN literature, including the MST, is that the input channels are clocked, i.e. the model is updated in discrete time. It is straightforward to extend the GNM model to the  continuous case (corresponding to eq. \ref{contmodel}). We found that training the model in continuous time yielded qualitatively the same results as the discrete time case. 
We demonstrate the feasibility of this interpretation by training a continuous-time version of the GNM to recognise two classes of patterns; see fig. \ref{fig:cont_example}.
The extension to the continuous case is interesting because then the GNM model with $\eta=0$ (eq. \ref{minmodelc}) can be interpreted as the description of a molecular species $V$ that decays with a rate of $\alpha$. The input $I$ is then mathematically equivalent to $N$ different input chemical species $I_i$, each decaying to $V$ with a rate of $w_i C$ and to a ``null-species'' with a rate of $(1-w_i)C$. In this case, the constant $C$ sets the time-scale of the decay and could be the same for all  ``pre-synaptic neurons''.  Having interpreted the GNM as a chemical system, we can then test its ability to recognise patterns by solving the differential equation \ref{minmodelc}. 
\par
\begin{figure}
\includegraphics[width=1\textwidth]{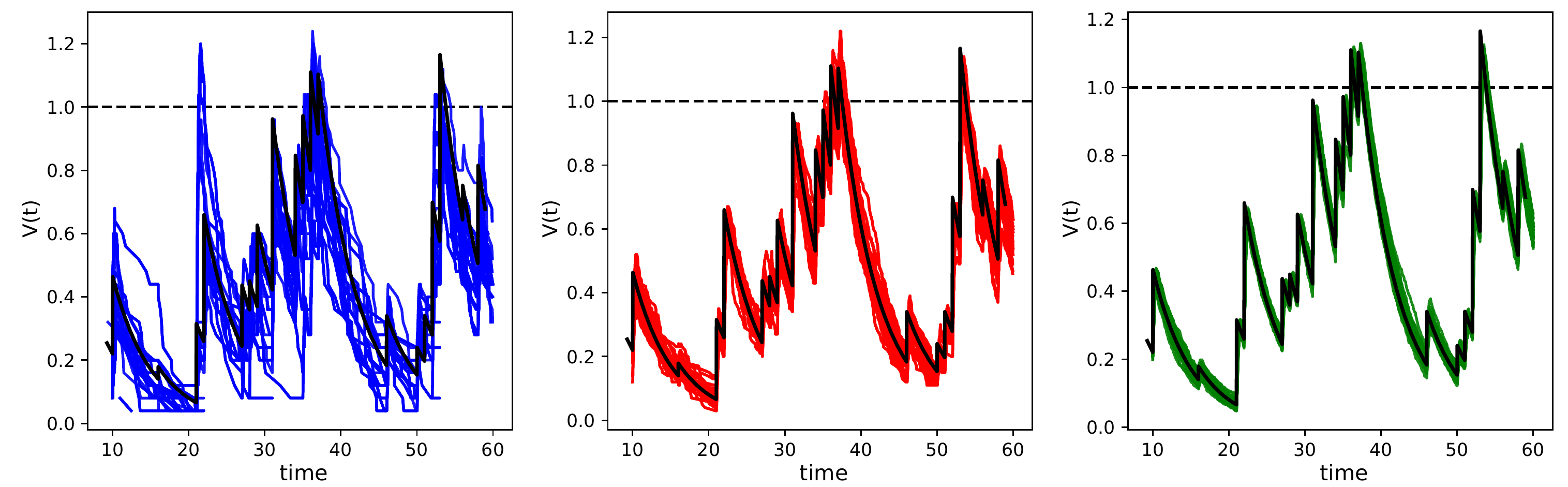}\label{CRN1}
\caption{ 
Membrane potential of a deterministic continuous-time GNM neuron trained to recognise two classes of patterns (black line), and equivalent stochastic chemical reaction network simulations with an input ``spike'' equivalent to 25, 100, and 500 molecules. For parameters $\eta=0$, $\alpha=0.2$, $\beta=0.3$.
The pre-synaptic spikes were encoded as instantaneous increase of corresponding pre-synaptic ``species'' by $N$, where $t^i_j$ is the time of the  $j$-th spiking event of the $i$-th pre-synaptic neuron, and $N={25,100,500}$ is  the number of particles that is added  to the pre-synaptic species $i$ at time $t_i^j$, $C$ was set to 10.  
}
\label{fig:CRNb}
\end{figure}

An underlying assumption of the differential equation models is that the number of molecules involved in the system is very large (technically infinite), such that $V$ can be described as a concentration. In any real system, the number of particles is finite. Indeed, in many biological information processing tasks there may only be a small number of particles involved in the computation. In this case, the system will exhibit noise around the exact solution of eq. \ref{minmodelc}; see fig. \ref{fig:CRNb} for a comparison of the stochastic and the deterministic solution. A concrete consequence of this is that the output of the neuron becomes stochastic with more or less frequent incorrect outputs, depending on the number of particles. 
\par

\section{Discussion}
\label{discussion}
In this contribution we probed the minimal ingredients necessary for neural computation in the context of multi-label classification of spatio-temporal patterns. We introduced the GNM which can solve multi-label classification tasks at least equally well as the MST, while being purely state-based. The model also has a ``conservation of membrane potential'' built in.  This does not preclude leakage of membrane potential, but prevents its creation out of nothing. In that sense, the model is physically plausible, which allows it to be interpreted in terms of concrete implementations (see below). 
\par
The dynamics of the GNM are simple and its parameters easily interpretable, which supports an intuitive understanding of what precisely it is that makes single neuron classification work. The parameter $\alpha$ can be interpreted as a memory. It  determines how much membrane potential is leaked between two update steps.  In the extreme case of $\alpha=1$, the neuron is reset during each time-step and has no memory of past inputs. In this limit the GNM is reduced to a standard rate-coding neuron. The performance of the GNM is substantially decreased, but some learning is still possible. For $\alpha=0$ the neuron integrates over all past events and never forgets. It is clear both from basic considerations but also from our simulation results that this latter limit does not allow the GNM to recognise patterns. In-between those two extremes there is an optimal value for the memory of the GNM. Figs. \ref{fig:fea1c} and \ref{optimal_a} suggest that the model is not particularly sensitive to the memory parameter, at least not for intermediate values. Interestingly, however, our simulations also suggest that  at the lower end of the parameter range, there is a critical value of $\alpha$ which separates almost perfect ability to learn from complete non-performance. 
\par
The second key parameter is the model choice parameter $\eta$. It controls the extent to which the neural dynamics is impacted by an internal threshold and hysteresis, in short how much ``spikiness'' the neural dynamics exhibits.  For $\eta=0$ the internal dynamics is a simple exponential decay with time constant $\alpha$. No thresholds are defined internally and there is no spiking whatsoever. Note, however, that  the use of the neuron still requires  an evaluation threshold $\vartheta_\textrm{R}$ to be set, in order to be able to interpret the membrane potential of the GNM as indicating the pattern class. 
\par
Increasing  $\eta$  introduces an additional behavioural threshold parameter $\vartheta_\textrm{B}$, which now {\em does} impact on the internal dynamics of the GNM. As the membrane potential nears  $\vartheta_\textrm{B}$  an additional decay term $R$ becomes relevant, such that after crossing the threshold the decay may be higher than before crossing the threshold. This endows the GNM with a spikiness and, most of all, with a time-limited memory of past spiking events 
\par
We found that model performance was consistently best  along an off-centre diagonal in the lower left quadrant of the heatmaps (fig.~\ref{fig:oneexp} \& \ref{manyexp}). Crucially, however, there is no consistent evidence for an optimal point along this diagonal. The conclusion to draw from this is that it is sufficient to consider the reduced parameter space corresponding to $\eta=0$ --- the case of no spikes. Put differently, there does not appear to be any benefit in spiking. 
\par
The existence of the optimal line provides some insights into the necessary ingredients for spiking networks in that it points to   the memory  parameter $\alpha$ as the main determinant of performance:   The optimal line, albeit not very well defined in our models, is the line of constant memory, because the factor $(1-\eta)$ effectively reduces the memory parameter $\alpha$. Therefore,  it appears from our  simulations that there is an optimal ``memory'' for the performance of the GNM which lies in-between the extreme and non-performing  cases of $\alpha=0$ and $\alpha=1$ corresponding to no forgetting and no memory at all. However, note that the model is not particularly sensitive to the precise value of $\alpha$, such that there is a range of values for which performance is good.  
\par
The conclusion that the GNM can perform with no spiking, opens an interesting perspective. For $\eta=0$ the GNM model eq. \ref{minmodelc} looks formally  like the time-evolution of the concentration $V$ of a molecular species subject to decay, i.e. $V \overset{\alpha}{\longrightarrow} \varnothing$ plus occasional instantaneous increases of $V$, i.e. $V \longrightarrow k V$.  Fundamentally, such a chemical system is an extremely simple system that can be implemented easily in  (wet) experiments. Yet, as we show, this simple system is sufficiently rich in its dynamics in order to perform the multi-label classification as well as the specialised MST model. All the chemical system retains in common with the MST is the temporal correlation of the input. This leads us to conjecture that this temporal autocorrelation is a crucial element  for multi-label classification of spatio-temporal patterns.
\par
This formal equivalence of GNM and chemical systems begs the questions whether or not there actually are man-made or naturally occurring chemical systems that recognise spatio-temporal patterns. An obvious place to look for such systems are biochemical-networks. It is conceivable that multi-label classification is exploited by gene regulatory networks to control gene expression by means of sequences of gene expression events.  
\par
Having shown that very simple systems  can perform multi-label classification,  it is instructive to compare the GNM and MST to another very simple model of a spiking neuron --- the LIF neuron.  The LIF neuron is different from both the GNM and the MST in that it has a hard reset following a spike and typically undergoes a refractory period. The refractory period, together with lateral inhibition, is a useful feature in the context of STDP learning \citep{Feldman2012, gerstnerhebb} which can be used to facilitate a {\em winner takes all} dynamics in multi-layer SNN networks, which in turn is important to prevent all post-synaptic neurons from learning the same parts of the input. Beyond that, it is unclear whether or not there is a computational benefit in the refractory period. 
\par
Our simulations showed (see fig. \ref{fig:LiF_GNM}) that the LIF has a comparable performance to the MST/GNM when classifying a single pattern, but its ability to learn drops for multi-pattern classification. This is understandable because the refractory period effectively shortens the time the LIF is able to react to incoming signals, thus making it hard for the neuron to activate several times during a limited period. 
Yet, as our simulations show (see fig. \ref{fig:LiF_GNM}),  the performance of the LIF neuron is worse even for a refractory period of length 0. Once the refractory period is removed, the only remaining difference between the LIF and the GNM is the hard-reset. Note that this hard reset effectively destroys the temporal autocorrelation of the membrane potential. Hence, the observation that the LIF neuron performs worse than the GNM supports further  our above conclusion that a balanced ``memory'' of the membrane potential is required for good performance on multi-label classification of spatio-temporal signals. This now raises the question whether biological neurons, which clearly do have a refractory period, are sub-optimal components. We do not believe that this conclusion can be made because brains operate in a different context from the restricted problem set that we considered here. Moreover, the refractory period in real neurons may well be a reflection of some resource limitations or physical constraints that we have not considered here, thus making a comparison invalid.
\par
Throughout this article, we evaluated the GNM assuming an aggregate label delayed feedback learning rule during training. This  training method is mainly motivated by its biological plausibility. In applications of the GNM/MST in the context of AI, biological plausibility is not a relevant criterion. We found, perhaps rather unsurprisingly, that dropping the requirement of aggregate label delayed feedback in favour of more immediate and information-rich feedback led to increased model performance (see section \ref{comparison}).
\par
Once we allow such direct error feedback, we can further extend it to backpropagation-based training methods in networks of GNMs.
We demonstrate the feasibility of this approach by solving the multi-label classification task using a layered network of GNMs with 10 hidden neurons; see fig. \ref{fig:BPa} and  \ref{fig:comp}. Deep learning with SNNs could lead to substantive benefits in terms of smaller models and more efficient hardware, if only it is possible to transfer established deep learning techniques to spiking architectures. We leave it to future research to establish whether the GNM or similar spiking architectures could indeed be a credible alternative for existing deep architectures.   
\section{Conclusions and Outlook}

G{\"u}tig's multi-spike tempotron is a powerful single neuron model that can classify spatio-temporal patterns into multiple classes. The model is also complicated to implement. Here, we showed that the much simpler GNM neuronal model can achieve the same performance as the MST. Our results indicate that the important feature of neuronal models is the temporal autocorrelation of the membrane potential, i.e. how quickly the neuron forgets about past inputs. We found that for intermediate values the model performance is maximised.
It remains an open question for future research whether this conclusion is specific to the particular task we considered, or whether the optimal memory emerges as {\em the} crucial parameter in all applications. If the power of SNNs is to be leveraged in practical AI applications, then it will be necessary to understand the minimal spiking neuron that is sufficient for a particular task so as to be able to build resource efficient systems.

\end{document}